\documentclass[conference]{IEEEtran}
\IEEEoverridecommandlockouts

\usepackage{cite}
\usepackage{amsmath,amssymb,amsfonts}
\usepackage{algorithmic}
\usepackage{graphicx}
\usepackage{textcomp}
\usepackage{xcolor}
\usepackage{booktabs}   
\usepackage{multirow}   
\usepackage{graphicx}  
\usepackage{makecell}  
\usepackage{comment}
\usepackage{mathtools}
\usepackage{stmaryrd}
\usepackage{bbm}
\usepackage{bm}
\def\BibTeX{{\rm B\kern-.05em{\sc i\kern-.025em b}\kern-.08em
    T\kern-.1667em\lower.7ex\hbox{E}\kern-.125emX}}

\begin{document}

\title{Polar-domain Segmentation of Peripapillary Atrophy and Optic Disc}

\author{
  \IEEEauthorblockN{
    Mengxian He\textsuperscript{1,*}
    \thanks{Mengxian He, Yunyun Sun, and Ziyue Gao
    contributed equally to this work and share first authorship.},
    Yunyun Sun\textsuperscript{2,*},
    Ziyue Gao\textsuperscript{1,*},
    Wengkei Lam\textsuperscript{1},
    Shunyi Zhang\textsuperscript{2}, and
    Wu Yuan\textsuperscript{1}
  }
  \IEEEauthorblockA{
    \textsuperscript{1}The Chinese University of Hong Kong,
    Hong Kong SAR, China\\
    Email: \{1155231596, 1155233115\}@link.cuhk.edu.hk,
    lamwengkei2004@gmail.com, wyuan@cuhk.edu.hk
  }
  \IEEEauthorblockA{
    \textsuperscript{2}Capital Medical University,
    Beijing, China\\
    Email: 2008sunshinsyy@163.com, 18519398200@163.com
  }
}

\maketitle

\section{Abstract}
Automatic segmentation of pepripapillary (PPA) and the optic disc (OD) from fundus photographs enables quantitative morphological analysis in myopic eyes. However, variations in OD-PPA geometry and the indistinct appearance of PPA complicate the delineation of their shared and outer boundaries. Motivated by the spatial arrangement of PPA around the OD, we reformulate their delineation by indexing the OD and PPA contours as two radial profiles in polar coordinates. Built upon this representation, we propose \textit{\textbf{UPolarSeg}}. To accommodate the polar geometric representation under non-uniform spatial sampling, \textit{\textbf{UPolarSeg}} equips U-shape skip connections with Radial-Angular Decoupling Module (RADM) to capture radial transitions and circumferential continuity respectively. A training-only auxiliary boundary decoder further promotes contour fidelity, supporting direct morphological measurements from the predicted segmentation masks. On internal and external cohorts, \textit{\textbf{UPolarSeg}} achieves PPA Dice scores of 89.59\% and 77.46\%, and OD Dice scores of 98.36\% and 97.63\%. For morphological quantification, the mean absolute errors are 2.4 and 6.7 pixels for maximum PPA width, and 0.017 and 0.036 for its ratio to the vertical OD diameter.

\begin{IEEEkeywords}
Fundus, Myopia, Optic Disc, Peripapillary Atrophy, Segmentation, Quantification.
\end{IEEEkeywords}
\section{Introduction}
Myopia is a leading cause of global visual impairment with effects beyond refractive errors. Its progression, characterized by continuous axial elongation and posterior pole remodeling, causes irreversible fundus alterations, particularly Peripapillary Atrophy (PPA)~\cite{ohno2021imi} and Optic Disc (OD) deformation, elevating the risk of blinding complications~\cite{he2026specf2m,sung2020parapapillary,liu2023quantitative}. Consequently, segmentation and quantification of PPA and OD as fundus biomarkers offers significant potential for cost-effective ocular disease screening and progression monitoring~\cite{liu2023quantitative,he2024ssl}.

Accurate OD and PPA segmentation poses both anatomical and visual challenges. First, substantial myopic OD variations lead to erratic shared and outer boundary trajectories. Second, the diffuse apperance and poor tissue contrast of PPA further degrade boundary visibility. Recent studies alleviate this difficulty by enhancing feature discrimination through multi-task learning~\cite{wei2023peripapillary} or frequency-domain refinement~\cite{wang2025frequency}; however, they largely overlook the underlying spatial geometry. Operating in the cartesian coordinates leaves the radial enclosing relationship between the OD and PPA implicit, placing an extensive learning burden on model to resolve rotational shape variations from limited annotation. This limitation motivates a coordinate mapping that explicitly re-indexes both PPA and OD boundaries with respect to the OD center.

\begin{figure}[t]
\centering
\includegraphics[width=1.0\linewidth]{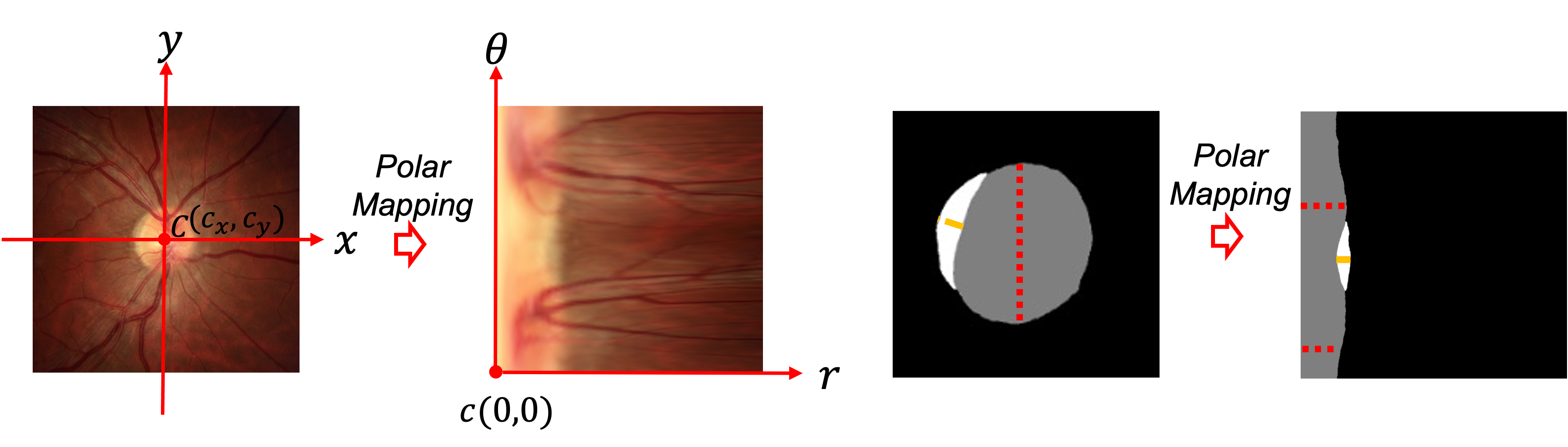}
\caption{Polar mapping. The vertical OD diameter and maximum PPA width can be derived from segmentation map.} \label{fig_2}
\end{figure}

Building on this geometric formulation, our key insight is to transform the OD and the PPA boundaries as two radial profiles parameterized by polar angle. Under this polar parameterization, the radial displacement between the profile trajectories of OD and PPA directly measures local PPA width across arbitrary angular meridians, establishing a closed-form correspondence between continuous boundary learning and morphological quantification (Fig. \ref{fig_2}). While polar representation and cartesian-polar hybrid network architectures have been explored for optic disc and cup segmentation~\cite{fu2018joint,liu2019ddnet}, a naive polar mapping introduces an intrinsic sampling distortion, as peripheral region span progressively larger physical distances per angular unit. Since standard convolutions assume spatially invariant sampling, feature representations must be conditioned on radial coordinates while decoupling radial transitions from circumferential continuity.

Grounded in the spatial-sampling dynamics, we propose \textit{\textbf{UPolarSeg}}, a polar-domain U-shaped architecture~\cite{ronneberger2015u} for OD and PPA segmentation. To account for the sampling distortion in polar grids, its Radial–Angular Decoupling Module (RADM) incorporates radial coordinate information into skip connections and separately models radial transitions and angular structural dependencies. We further introduce an auxiliary decoder to supervise sophisticated boundary learning. From the predicted segmentation masks, critical clinical metrics including the peak maximum PPA width (PMW) and its ratio to the vertical OD diameter (PVDR)~\cite{moon2021relationship} are extracted directly via geometric indexing. Our main contributions are threefold: 
\begin{itemize}
    \item \textbf{Geometric Formulation and Sampling Analysis:} we formulate OD and PPA delineation as angle-indexed radial profiles and reveal how intrinsic spatial distortion degrades polar mapping, providing a geometric rationale for \textit{\textbf{UPolarSeg}}.
    \item  \textbf{Polar-Decoupled Network:} We design \textit{\textbf{UPolarSeg}}, equipped with the Radial-Angular Decoupling Module (RADM) and training-only contour supervision, explicitly tailoring feature extraction to polar geometry.
    \item \textbf{Comprehensive Clinical Validation:} Cross-center evaluations demonstrate that \textit{\textbf{UPolarSeg}} achieves highest PPA Dice scores of 89.59\% and 77.46\%, respectively. On the external validation set, UPolarSeg cuts the mean absolute errors of PMW and PVDR by 17.3\% and 16.3\%, demonstrating strong cross-domain reliability.
\end{itemize}

\section{Preliminary Study}
\begin{figure}[t]
\centering
\includegraphics[width=0.5\textwidth]{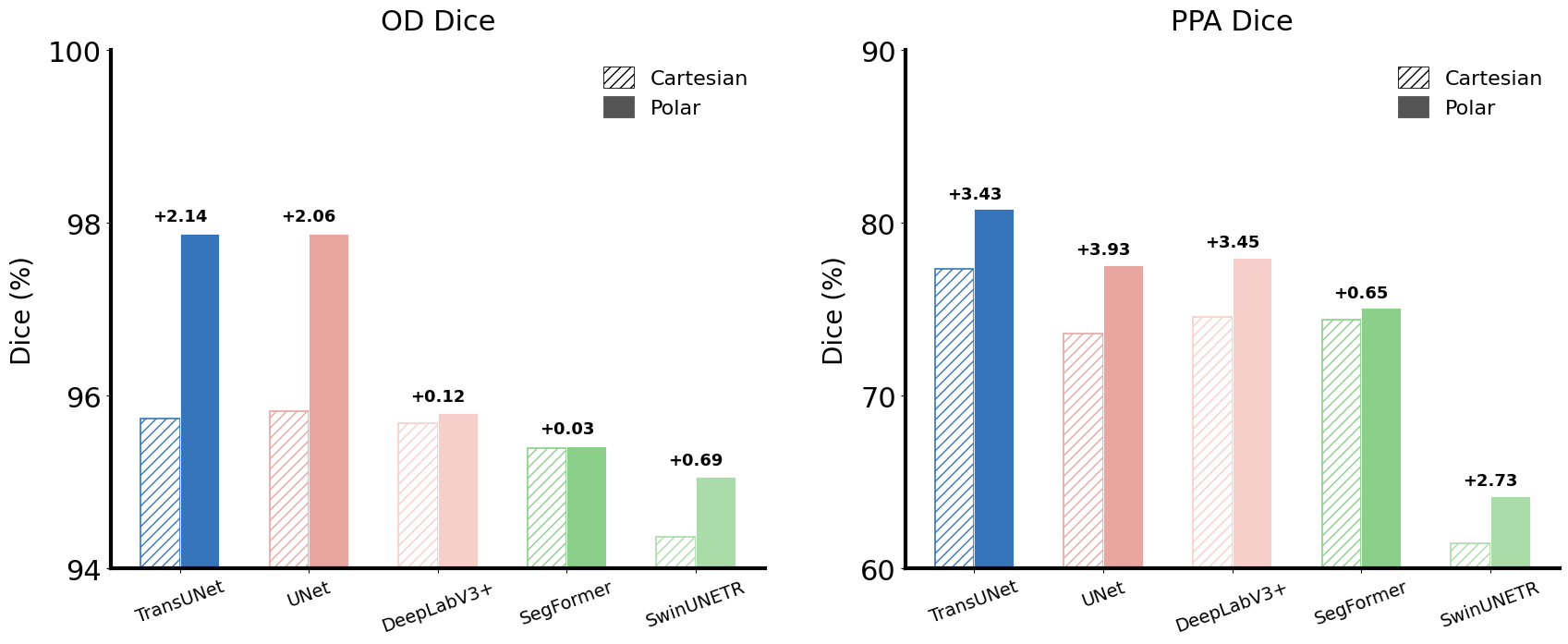}
\caption{Effect of polar representation on CNN, ViT, and hybrid backbones. Dice scores are averaged two cohorts.}
\label{fig_pre_experiments}
\end{figure}

We examine the geometric properties of polar representation and assess its effect on OD/PPA segmentation (Fig. \ref{fig_pre_experiments}).

\subsection{Radial Boundary Representation}
Given a fundus image $I$ and an OD center $c=(c_x,c_y)$,
the polar transform $P_c$ produces
\begin{equation}
    I_p(\theta,r) = I(c_x+r\cos\theta,\;c_y+r\sin\theta),
\label{eq:polar_transform}
\end{equation}
where $\theta\in[0,2\pi)$ and $r\in[0,R_{\max}]$. Under an approximation about $c$, the OD and PPA region, denoted OP, have radial boundaries
\begin{equation}
\Gamma_k^p = \{(\theta,R_k(\theta))\}, \quad k\in\{\mathrm{OD},\mathrm{OP}\}.
\label{eq:radial_profiles}
\end{equation}
This expresses each contour as a radial position at each angle. The difference $R_{\mathrm{OP}}(\theta)-R_{\mathrm{OD}}(\theta)$ gives the local radial PPA width. These profiles provide a common description for boundary supervision and morphological measurements.

\subsection{Geometric Properties and Empirical Evidence}
For a rotation $\mathcal{R}_{\phi}$ and a scaling $\mathcal{S}_{s}$ about the same center $c$, the continuous polar mapping satisfies
\begin{equation}
    \begin{aligned}
    P_c[\mathcal{R}_{\phi}I](\theta,r)
    &= I_p(\theta-\phi,r),\\
    P_c[\mathcal{S}_{s}I](\theta,r)
    &= I_p(\theta,r/s).
    \end{aligned}
\label{eq:polar_geometry}
\end{equation}
Thus, rotation becomes angular translation, while scaling becomes radial dilation. The angular correspondence motivates convolutional modeling of circumferential continuity. We next assess the representation empirically across five backbones, using the same architecture and training
settings within each cartesian-polar pair. As shown in Fig.~2, Dice scores improve by 0.65-3.93\% for PPA. These improvements support the use of polar representation across the evaluated backbones.

\subsection{Nonuniform Spatial Sampling}
Uniform polar sampling corresponds to nonuniform spacing in the cartesian coordinates:
\begin{equation}
    ds^2=dr^2+r^2d\theta^2, \qquad \Delta s_{\theta}(r)=r\Delta\theta.
\label{eq:polar_sampling}
\end{equation}
A fixed angular interval therefore spans a larger arc length at a larger radius, whereas radial spacing remains constant. This motivates incorporating radial position and separately refining radial transitions and circumferential continuity, as implemented by RADM in UPolarSeg.
\section{Methodology}
\begin{figure*}[t]
\centering
\includegraphics[width=1.0\linewidth]{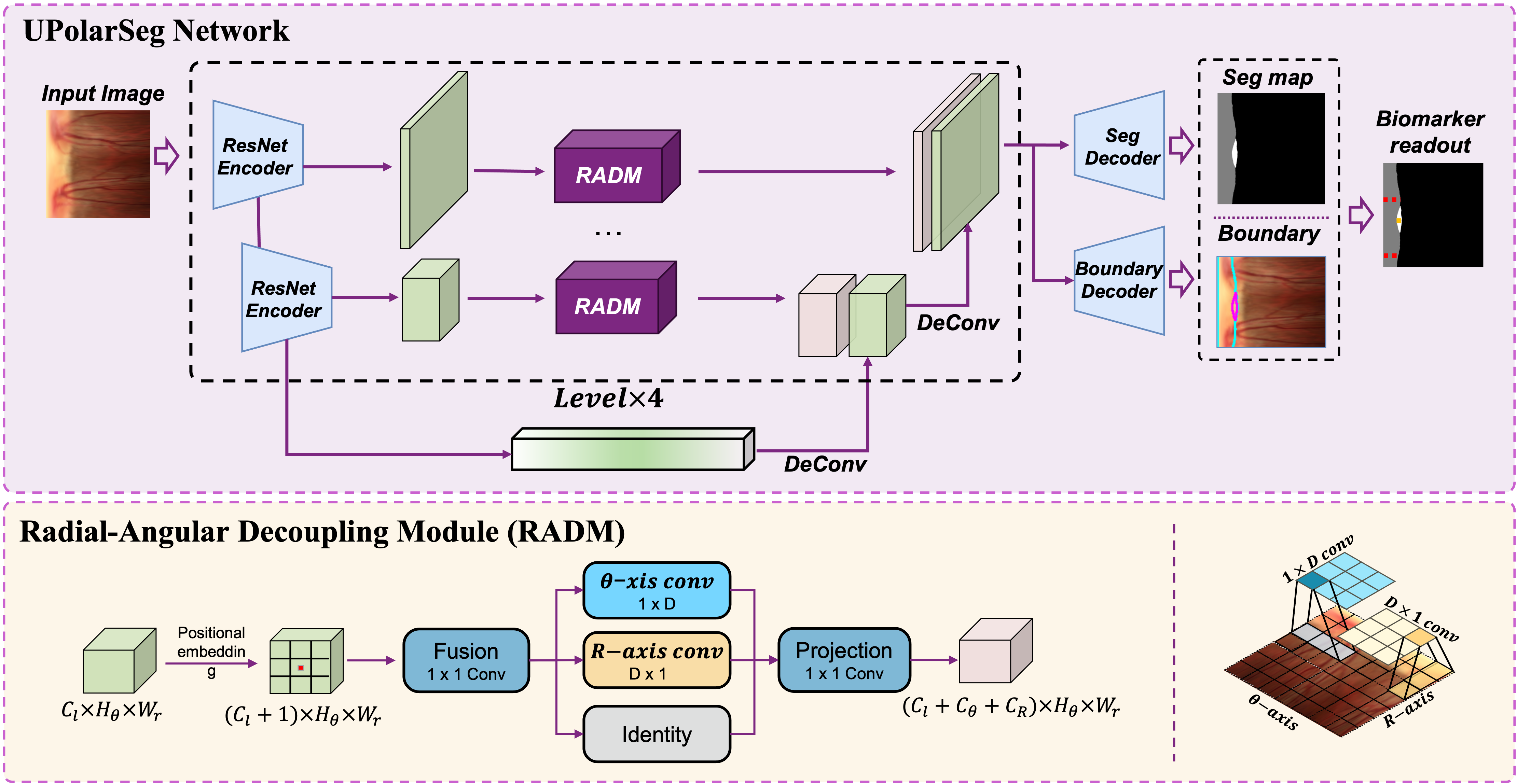}
\caption{Overall framework of UPolarSeg. UPolarSeg adopts a ResNet-based U-shape backbone enhanced by Radial-Angular-Decoupling modules (RADM) and an auxiliary boundary decoder.
} \label{fig:method}
\end{figure*}

\subsection{Overview}

UPolarSeg comprises OD localization, polar-domain segmentation, and morphological quantification (Fig. \ref{fig:method}). A coarse OD mask obtained by zero-shot RetSAM determines
the crop center. We extract a $1024\times1024$ region of interest, resize it to $512\times512$, and apply the polar mapping in Section III respect to the corresponding OD center. During training, annotations undergo the same spatial mapping. A ResNet-based U-shaped backbone predicts three semantic classes: background, OD, and PPA. RADM refines the skip features, while an auxiliary boundary decoder provides additional supervision during training.

Formally, the encoder-decoder backbone of UPolarSeg extracts multi-scale features from $I_p$. The final high-resolution decoder feature $D_0$ is shared by a primary semantic segmentation head $H_{\mathrm{seg}}$ and an auxiliary boundary head $H_{\mathrm{bd}}$:
\begin{equation}
    \hat{Z}^p = H_{\mathrm{seg}}(D_0), \quad \hat{B}^p = H_{\mathrm{bd}}(D_0),
\end{equation}
where $\hat{Z}^p$ and $\hat{B}^p$ denote the semantic and boundary logits, respectively. The final prediction is obtained via $\hat{Y}^p = \arg\max_m \mathrm{Softmax}(\hat{Z}^p)_m$, which assigns each pixel to background, OD, or PPA. During inference, $H_{\mathrm{bd}}$ is discarded, which make auxiliary geometric supervision cost-free.

\subsubsection{Radial-Angular-Decoupling Module}
\label{sec:radm}
Following the sampling analysis in Section III, RADM incorporates radial position and processes the two polar axes separately. For a skip feature $S_l$ with spatial dimensions $H_\theta^l\times W_r^l$, we append a normalized radial coordinate channel, $\rho_l(i,j)=j/(W_r^l-1)$, broadcast along the angular axis. The refined skip feature is computed as
\begin{equation}
    \begin{aligned}
    Z_l
    &= \operatorname{Conv}_{1\times1}
       \bigl([S_l,\rho_l]\bigr),\\
    \Delta_l^r
    &= \operatorname{Conv}_{1\times1}^{r}
       \bigl(\operatorname{DWConv}_{1\times k_r}(Z_l)\bigr),\\
    \Delta_l^\theta
    &= \operatorname{Conv}_{1\times1}^{\theta}
       \bigl(\operatorname{DWConv}_{k_\theta\times1}(Z_l)\bigr),\\
    S_l'
    &= [S_l,\Delta_l^r,\Delta_l^\theta],
    \end{aligned}
\label{eq:radm}
\end{equation}
where $[\cdot]$ denotes channel concatenation. The radial branch models tissue transitions across radii, while the angular branch models circumferential continuity and uses circular padding along $\theta$. The augmented feature $S_l'$ replaces $S_l$ in the corresponding skip connection.

\subsubsection{Auxiliary Boundary Supervision}
The semantic and auxiliary boundary heads share the final decoder feature $D_0$. The semantic head uses softmax to predict three classes, while the boundary head uses sigmoid to predict the OD and PPA boundaries. Given the polar label map $Y^p$, we define
$M_{\mathrm{OD}}=\mathbf{1}[Y^p=\mathrm{OD}]$ and
$M_{\mathrm{OP}}=
\mathbf{1}[Y^p\in\{\mathrm{OD},\mathrm{PPA}\}]$.
For $t\in\mathcal{T}=\{\mathrm{OD},\mathrm{OP}\}$,
the outermost foreground index and its boundary target are
\begin{equation}
    \begin{aligned}
    J_t(\theta)
    &= \max\{j \mid M_t(\theta,j)=1\},\\
    B_t(\theta,j)
    &= \mathbf{1}\bigl[|j-J_t(\theta)|\leq\tau\bigr],
    \end{aligned}
\label{eq:boundary_targets}
\end{equation}
where $\tau$ is the boundary-band half-width in radial bins.

The segmentation loss combines multiclass cross-entropy and soft Dice. For each boundary channel, binary cross-entropy and soft Dice are evaluated between the sigmoid prediction and
its target band:
\begin{equation}
    \begin{aligned}
    \mathcal{L}_{\mathrm{bd}}
    &= \frac{1}{2}\sum_{t\in\mathcal{T}}
       \left(
       \mathcal{L}_{\mathrm{BCE}}^t
       + \mathcal{L}_{\mathrm{Dice}}^t
       \right),\\
    \mathcal{L}
    &= \mathcal{L}_{\mathrm{CE}}
       + \mathcal{L}_{\mathrm{Dice}}
       + \lambda_{\mathrm{bd}}\mathcal{L}_{\mathrm{bd}}.
    \end{aligned}
\label{eq:training_objective}
\end{equation}
The auxiliary boundary branch is discarded at inference.

\subsection{Morphological Quantification}
At inference, the predicted semantic mask defines $\hat M_{\mathrm{OD}}$ and $\hat M_{\mathrm{OP}}$. Applying the outermost-index rule in Eq.~\eqref{eq:boundary_targets} gives
$\hat J_t(\theta)$ and the radial profiles $\hat R_t(\theta)=r_{\hat J_t(\theta)}$,
where $r_j$ is the radius assigned to bin $j$ by the polar mapping. With $\theta=0$ along the image horizontal axis, the OD diameters and PPA measurements are
\begin{equation}
    \begin{aligned}
    D_H
    &= \hat R_{\mathrm{OD}}(0)
       + \hat R_{\mathrm{OD}}(\pi),\\
    D_V
    &= \hat R_{\mathrm{OD}}(\pi/2)
       + \hat R_{\mathrm{OD}}(3\pi/2),\\
    \mathrm{PMW}
    &= \max_\theta
       \bigl[
       \hat R_{\mathrm{OP}}(\theta)
       - \hat R_{\mathrm{OD}}(\theta)
       \bigr]_+,\\
    \mathrm{HVDR}
    &= D_H/D_V,
    \qquad
    \mathrm{PVDR} = \mathrm{PMW}/D_V,
    \end{aligned}
\label{eq:morphological_measurements}
\end{equation}
where $[x]_+=\max(x,0)$.
\section{Experiment}
\begin{table*}[t!]
\centering
\caption{Main comparison of segmentation accuracy, quantification error, and model size under the polar representation.}
\label{tab:main_results}
\small
\renewcommand{\arraystretch}{1.15}
\setlength{\tabcolsep}{3.5pt}
\begin{tabular*}{\textwidth}{@{\extracolsep{\fill}}l c cc ccc cc ccc@{}}
\toprule
\multirow{3}{*}{\textbf{Method}}
& \multirow{3}{*}{\textbf{$\text{Params}$(M)}}
& \multicolumn{5}{c}{\textbf{Internal Cohort}}
& \multicolumn{5}{c}{\textbf{External Cohort}} \\
\cmidrule(lr){3-7} \cmidrule(lr){8-12}
&
& \multicolumn{2}{c}{\textbf{Dice (\%) $\uparrow$}}
& \multicolumn{3}{c}{\textbf{MAE $\downarrow$}}
& \multicolumn{2}{c}{\textbf{Dice (\%) $\uparrow$}}
& \multicolumn{3}{c}{\textbf{MAE $\downarrow$}} \\
\cmidrule(lr){3-4} \cmidrule(lr){5-7}
\cmidrule(lr){8-9} \cmidrule(lr){10-12}
&
& \textbf{OD}
& \textbf{PPA}
& \textbf{HVDR}
& \textbf{PMW}
& \textbf{PVDR}
& \textbf{OD}
& \textbf{PPA}
& \textbf{HVDR}
& \textbf{PMW}
& \textbf{PVDR} \\
\midrule

UNet
& 24.4
& 98.04 & 88.48 & 0.043 & 3.2 & 0.019
& \textbf{97.78} & 70.61 & 0.037 & 13.4 & 0.068 \\

SwinUNETR
& 25.1
& 96.16 & 86.33 & 0.044 & 4.0 & 0.023
& 94.36 & 50.14 & \textbf{0.033} & 14.1 & 0.073 \\

DeepLabV3+
& 26.7
& 96.51 & 88.74 & 0.038 & 3.1 & 0.020
& 95.34 & 71.16 & 0.040 & 8.1 & 0.043 \\

SegFormer
& 27.3
& 96.36 & 87.99 & 0.038 & 3.3 & 0.020
& 94.82 & 66.87 & 0.036 & 12.6 & 0.065 \\

TransUNet
& 105.9
& 98.21 & 88.52 & 0.034 & 2.5 & \textbf{0.015}
& 97.67 & 75.88 & 0.042 & 8.4 & 0.043 \\

RetSAM
& 767.2
& 95.46 & 83.23 & 0.045 & 4.3 & 0.023
& 95.09 & 64.11 & 0.059 & 11.8 & 0.062 \\

\textbf{UPolarSeg}
& 24.9
& \textbf{98.36} & \textbf{89.59} & \textbf{0.028} & \textbf{2.4} & 0.017
& 97.63 & \textbf{77.46} & 0.048 & \textbf{6.7} & \textbf{0.036} \\


\bottomrule
\end{tabular*}

\vspace{0.4em}
\begin{minipage}{0.98\textwidth}
\footnotesize
\textit{Note.}
PMW MAE is measured in pixels. All methods except RetSAM are trained and evaluated under the polar representation. 
\end{minipage}
\end{table*}

\subsection{Dataset}
We evaluated UPolarSeg on two independent cohorts, ethics committee approval was obtained from the Institutional Review Board of Beijing Tongren Hospital, Capital Medical University. All participating students gave their informed consent before participating in the study, according
to the tenets of the Declaration of Helsinki. The internal cohort from Beijing comprises 388 myopic fundus images from 114 patients (patient-level splits of train/val/test: 307/39/42). The external cohort from Tibet consists of 66 myopic images from 51 patients, used only for out-of-distribution evaluation. Image resolutions are 2592×1728 (internal) and 2032×1934 (external). All images were preprocessed by localizing the OD center with RetSAM, cropping a 1024×1024 OD-centered region, and resizing to 512×512 for network input.

\subsection{Implementation details and evaluation metric }
Polar-domain learning mapped crops to a polar grid (radius 362 px). Training used AdamW for 150 epochs (initial learning rate: $1 \times 10^{-4}$, weight decay: $1 \times 10^{-4}$), cosine annealing to $1 \times 10^{-6}$, and early stopping (patience 30). The architecture was frozen before external evaluation. Segmentation evaluation used Dice. Quantification evaluation used Mean Absolute Error (MAE) of HVDR, PMW, and PVDR. Computational efficiency used the number of inference parameters ($\text{Params}$). 

\subsection{OD localization reliability}
\begin{table}[t]
\centering
\caption{Coarse segmentation results and bounding box containment rates.}
\label{tab:coarse_combined}
\resizebox{\columnwidth}{!}{ 
\begin{tabular}{lcccc}
\toprule
\multirow{2}{*}{\textbf{Cohort}} 
& \multirow{2}{*}{\textbf{OD Dice (\%)}} 
& \multicolumn{2}{c}{\textbf{Localization Error (px)}} 
& \multirow{2}{*}{\textbf{Containment Rate}} \\
\cmidrule(r){3-4}
& & \textbf{Center} & \textbf{Worst-case} & \\
\midrule
Beijing & 94.67 & $5.9 \pm 6.2$  & 72.1 & 100.0\% \\
Tibet   & 94.41 & $10.4 \pm 3.8$ & 23.7  & 100.0\% \\
\bottomrule
\end{tabular}
}
\end{table}
Table~\ref{tab:coarse_combined} shows that zero-shot RetSAM yielded mean OD-center errors of $5.9\pm6.2$ pixels internally and $10.4\pm3.8$ pixels externally, with 100\% crop containment in both cohorts. These results support automatic ROI extraction for the evaluated images.

\subsection{Main Results and Comparative Analysis}
Table~\ref{tab:main_results} compares network designs under the polar mapping, with fine-tuned RetSAM serving as a cartesian reference. Among the baselines, DeepLabV3+ achieves the highest internal PPA Dice, whereas the hybrid TransUNet performs best externally. SegFormer and SwinUNETR do not surpass the convolutional baselines in PPA Dice under this protocol. UPolarSeg improves PPA Dice over the best comparator by 0.85\% and 1.58\% on the internal and external cohorts, respectively. Compared with the polar U-Net, it improves external PPA Dice from 70.61\% to 77.46\% with 0.5M additional segmentation parameters. This comparison supports the combined value of feature refinement and boundary supervision beyond coordinate mapping alone.

Segmentation and measurement rankings are not identical: TransUNet has higher external PPA Dice than DeepLabV3+, but a higher PMW MAE (8.4 versus 8.1 pixels). UPolarSeg reduces external PMW and PVDR MAEs to 6.7 pixels and 0.036, corresponding to reductions of 17.3\% and
16.3\% relative to the best comparator for each metric. Its internal PVDR and external HVDR errors are not the lowest, indicating that the advantages are concentrated in PPA segmentation and external PPA-related measurements.

\subsection{Ablation Study}
\begin{table}[h]
\centering
\caption{Ablation study under the polar representation.}
\label{tab:ablation}
\scriptsize
\setlength{\tabcolsep}{1.7pt}
\renewcommand{\arraystretch}{1.08}
\begin{tabular*}{\columnwidth}{@{\extracolsep{\fill}}c ccc cc ccc@{}}
\toprule
\multirow{2}{*}{\textbf{Coh.}}
& \multicolumn{3}{c}{\textbf{Modules}}
& \multicolumn{2}{c}{\textbf{Dice (\%) $\uparrow$}}
& \multicolumn{3}{c}{\textbf{MAE $\downarrow$}} \\
\cmidrule(lr){2-4} \cmidrule(lr){5-6} \cmidrule(lr){7-9}
& \textbf{RAD$_r$}
& \textbf{RAD$_\theta$}
& \textbf{BD}
& \textbf{OD}
& \textbf{PPA}
& \textbf{HVDR}
& \textbf{PMW}
& \textbf{PVDR} \\
\midrule

\multirow{6}{*}{Int.}
& -- & -- & --
& 98.04 & 88.48 & 0.043 & 3.2 & 0.019 \\

& \checkmark & -- & --
& 98.23 & 88.81 & 0.031 & 2.7 & 0.018 \\

& -- & \checkmark & --
& 98.28 & 88.94 & 0.033 & 2.8 & 0.019 \\

& \checkmark & \checkmark & --
& 98.24 & 89.30 & 0.035 & 2.7 & 0.020 \\

& -- & -- & \checkmark
& 98.14 & 88.60 & 0.030 & \textbf{2.2}& 0.017 \\

& \checkmark & \checkmark & \checkmark
& \textbf{98.36} & \textbf{89.59}
& \textbf{0.028} & 2.4 & \textbf{0.017} \\

\midrule

\multirow{6}{*}{Ext.}
& -- & -- & --
& \textbf{97.78} & 70.61 & \textbf{0.037} & 13.4 & 0.068 \\

& \checkmark & -- & --
& 97.71 & 72.50 & 0.047 & 10.5 & 0.054 \\

& -- & \checkmark & --
& 97.72 & 75.00 & 0.051 & 8.0 & 0.043 \\

& \checkmark & \checkmark & --
& 97.65 & 75.31 & 0.043 & 8.8 & 0.046 \\

& -- & -- & \checkmark
& 97.73 & 75.11 & 0.038 & 8.3 & 0.043 \\

& \checkmark & \checkmark & \checkmark
& 97.63 & \textbf{77.46}
& 0.048 & \textbf{6.7} & \textbf{0.036} \\

\bottomrule
\end{tabular*}

\vspace{0.3em}
\begin{minipage}{\columnwidth}
\scriptsize
\textit{Note.}
RAD$_r$ and RAD$_\theta$ denote radius- and angular-axis refinement, respectively.
BD denotes the auxiliary boundary decoder.
\end{minipage}
\end{table}

Table~\ref{tab:ablation} evaluates directional refinement and the auxiliary boundary decoder (BD) using the same polar U-Net backbone. Angular refinement increases external PPA Dice by
4.39\%, compared with 1.89 for radial refinement. This larger observed gain supports incorporating circumferential context alongside radial boundary localization. BD alone lowers internal PMW MAE from 3.2 to 2.2 pixels, while PPA Dice increases by only 0.12 percentage points. Because PMW depends on the maximum radial boundary separation, improved width estimates need not be accompanied by a large change in regional overlap.

Adding BD to the joint radial-angular model improves external PPA Dice from 75.31\% to 77.46\% and reduces PMW/PVDR MAEs from 8.8/0.046 to 6.7/0.036. The full configuration therefore provides the best external PPA segmentation and width measurements among the evaluated variants. The effects are nevertheless metric-dependent: directional refinement has a higher external PMW error than angular refinement alone, while BD alone gives the lowest internal PMW MAE.

\section{limitation}
Current evaluation of this work was limited to 42 internal and 66 external test images, warranting validation in larger and more diverse cohorts. Errors in PPA-negative eyes have not been systematically characterized, particularly false-positive segmentation and spurious PPA-width estimates. The training objective supervises segmentation and boundaries but does not explicitly minimize errors in the derived morphological measurements; joint optimization
of segmentation and measurement accuracy warrants further investigation. Finally, the clinical utility of the extracted measurements remains insufficiently explored and requires further validation through independent clinical assessment and longitudinal study.
\section{Conclusion}
We presented UPolarSeg for OD/PPA segmentation and morphological quantification from fundus photographs. The framework combines OD-centered polar representation, RADM in skip connections, and training-only auxiliary boundary supervision. UPolarSeg achieved PPA Dice scores of 89.59\% and 77.46\% on the internal and external cohorts, respectively, and reduced external PMW and PVDR MAEs by 17.3\% and 16.3\% relative to the best comparator for each metric. These findings support adapting feature processing and supervision to polar geometry for PPA segmentation and
direct width quantification from predicted masks.

\bibliographystyle{IEEEbib}
\bibliography{strings,mybib}

\end{document}